\documentclass[10pt,twocolumn,letterpaper]{article}

\usepackage{wacv}
\usepackage{times}
\usepackage{epsfig}
\usepackage{graphicx}
\usepackage{amsmath}
\usepackage{amssymb}
\usepackage{subcaption}
\usepackage{booktabs}
\usepackage{tabularx}
\usepackage{multirow}
\usepackage{enumitem}
\DeclareMathOperator*{\argmax}{arg\,max}

\usepackage[pagebackref=true,breaklinks=true,letterpaper=true,colorlinks,bookmarks=false]{hyperref}

\wacvfinalcopy % *** Uncomment this line for the final submission

\def\wacvPaperID{219} % *** Enter the wacv Paper ID here

\ifwacvfinal\fi
\makeatletter
\let\origparagraph\paragraph
\renewcommand\paragraph{\@ifstar{\starparagraph}{\nostarparagraph}}
\newcommand\nostarparagraph[1]
{\origparagraph{#1}\paragraphpostlude}
\newcommand\starparagraph[1]
{\paragraphprelude\origparagraph*{#1}\paragraphpostlude}
\newcommand\paragraphprelude{%
  \vspace{-14pt}% STUFF TO DO PRIOR TO THE SECTION HEADING
}
\newcommand\paragraphpostlude{%
}

\begin{document}

%%%%%%%%% TITLE
\title{Attention Mechanisms for Object Recognition with Event-Based Cameras}

% Authors at the same institution
%\author{First Author \hspace{2cm} Second Author \\
%Institution1\\
%{\tt\small firstauthor@i1.org}
%}
% Authors at different institutions
\author{Marco Cannici \hspace{20pt} 
Marco Ciccone \hspace{20pt} 
Andrea Romanoni\hspace{20pt} 
Matteo Matteucci\\
Politecnico di Milano, Italy\\
{\tt\small \{marco.cannici,marco.ciccone,andrea.romanoni,matteo.matteucci\}@polimi.it}
% \and
% Marco Ciccone \\
% Politecnico di Milano, Italy\\
% {\tt\small marco.ciccone@polimi.it}
% \and
% Andrea Romanoni \\
% Politecnico di Milano, Italy\\
% {\tt\small andrea.romanoni@polimi.it}
% \and
% Matteo Matteucci \\
% Politecnico di Milano, Italy\\
% {\tt\small matteo.matteucci@polimi.it}
}

\maketitle
\ifwacvfinal\thispagestyle{empty}\fi

%%%%%%%%% ABSTRACT
\begin{abstract}
Event-based cameras are neuromorphic sensors capable of efficiently encoding visual information in the form of sparse sequences of events. Being biologically inspired, they are commonly used to exploit some of the computational and power consumption benefits of biological vision. In this paper we focus on a specific feature of vision: visual attention. We propose two attentive models for event based vision: an algorithm that tracks events activity within the field of view to locate regions of interest and a fully-differentiable attention procedure based on DRAW neural model. We highlight the strengths and weaknesses of the proposed methods on four datasets, the Shifted N-MNIST, Shifted MNIST-DVS, CIFAR10-DVS and N-Caltech101 collections, using the Phased LSTM recognition network as a baseline reference model obtaining improvements in terms of both translation and scale invariance.
\end{abstract}

\section{Introduction}

Convolutional neural networks (CNNs) are currently the state of the art in a variety of challenging computer vision tasks that involve the extraction of visual features. These include, among the others, image classification~\cite{xie2017aggregated, He2015Dec, Szegedy2016Feb}, object detection~\cite{Ren2015Jun, Redmon2017, Kang2018, Lin2017} as well as semantic scene labeling~\cite{Lin2017refinenet, Chen2018Apr, Lin2018}. When video sequences captured by conventional frame-based cameras are considered, CNNs great performance in terms of accuracy, however, is achieved at the cost of a high computational and time complexity.
Indeed even while capturing static scenes, these devices output a stream of mostly identical frames, requiring CNNs to process the same redundant data several times.

On the other hand, primates are able to achieve remarkable results in most vision tasks while using a fraction of energy and computational power with respect to their artificial counterparts. As an attempt to reproduce the benefits of biological vision, research is now focusing on developing vision systems based on neuromorphic, or event-based, cameras, \ie, a type of sensors that tries to emulate the functioning of biological retinas. Unlike conventional cameras, these devices output sequences of asynchronous events that efficiently encode pixel-level brightness changes caused by objects moving inside the scene. The result is a sensor able to produce a stream of events $e = \langle x, y, ts \rangle$ indicating the time instant $ts$, the position $(x, y)$ and the polarity $p \in \{-1, 1\}$  of every change detected inside the scene.

A key characteristic of biological vision systems is their ability to selectively focus their attention on the salient portions of the scene, drastically reducing the amount of information that needs to be processed. Selective attention mechanisms that mimic this behavior are nowadays widely adopted in several vision tasks, like for instance in image and video captioning \cite{Xu2015, Chen2017July, gao2017video}, image generation \cite{Gregor2015Feb}, object recognition \cite{Mnih2014Jun, wang2017residual} and person re-identification \cite{Song2018mask}. A similar effort has been made in the design of attention mechanisms able to directly process event-based information produced by neuromorphic cameras \cite{Sonnleithner2011Mar, Rea2013Dec}. 
These devices are indeed inherently able to detect relevant portions of the field of view as they emit events only when something changes. Events encode important information regarding the objects contained inside the scene and can thus be used to precisely locate them in space and time.

These neuromorphic systems make often use of Spiking Neural Networks (SNNs) \cite{Maass1997Dec}, a type of artificial neural networks based on units that communicate with each other through spikes and perform computation only when and where needed. However, a big limitation of these models is that they are not differentiable. When multiple processing layers are involved, this makes the training procedure much more complex than the back-propagation algorithm used in conventional neural networks. For this reason, another approach adopted in literature makes use of conventional convolutional or recurrent networks properly adapted to handle event based information \cite{Perez-Carrasco2013Nov, Neil2016Oct, Cannici2018May}. Despite being easier to train, however, such networks usually require the scene to be reconstructed as a sequence of frames, thus potentially ignoring all the advantages of the events encoding.

In this paper we focus on enhancing conventional architectures by designing attention mechanisms that can be used to make these networks focus only on relevant instants of events recordings and only on the salient portions of frames, limiting the increased data redundancy caused by the frame integration process.

The main contributions of this paper are:
\vspace{-5pt}
\begin{itemize}[noitemsep]
    \item An algorithmic attention mechanism which monitors the events activity within the scene to extract patches from reconstructed frames (Section \ref{sec:patch_extractor_network;chap:models}).
    \item An adaptation of the popular DRAW~\cite{Gregor2015Feb} attention mechanism for image classification able to recognize objects within reconstructed frames (Section \ref{sec:draw_patch_network;chap:models}).
    \item An event-based variant of the previous network which directly uses events to locate the relevant portions of the frame (Section \ref{sec:draw_event_network;chap:models}).
\end{itemize}

\section{Background}
This section presents three basic tools adopted to design the attention mechanisms proposed in this paper: the Phased LSTM Network, the DRAW attention mechanism and the Leaky Frame Integrator.

\vspace{-10pt}
\paragraph{Phased LSTM Recognition Network} \label{sec:phaselstm}

\begin{figure}
	\centering
	\includegraphics[width=0.85\linewidth]{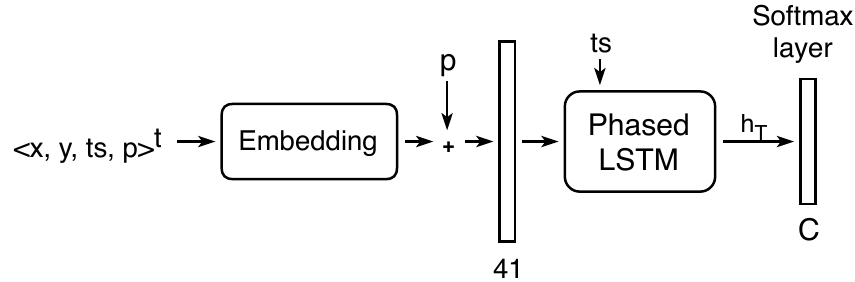}
    \vspace{-0.4cm}
	\caption{Phased LSTM recognition network.}
	\label{fig:phasedlstm_eventbased}
    \vspace{-0.4cm}
\end{figure}

The Phased LSTM recognition network \cite{Neil2016Oct} is a simple architecture for object classification with event-based cameras. It is based on Phased LSTM (pLSTM) cells, a variant of the vanilla LSTM which makes use of a \emph{time gate} to learn the time scales of incoming events, and uses of a word embedding layer to extract relevant features from a stream of events. Its structure is depicted in Figure \ref{fig:phasedlstm_eventbased}.

Despite achieving good results on simple datasets, however, the network lacks in the ability to extract general features as its embedding layer is only able to learn simple mappings between coordinates and learned sets of features. This results in a model with poor translation and scale invariance properties. This paper focuses on improving its performance in conditions where objects may appear with multiple scales and in different portions of the field of view.

\vspace{-10pt}
\paragraph{DRAW Selective Attention}
\label{sec:draw}

The \emph{Deep Recurrent Attentive Writer (DRAW)} \cite{Gregor2015Feb} is a network for image generation that makes use of a novel fully-differentiable procedure to focus attention on the salient portion of a frame. Its core components are a \emph{recurrent neural network} (RNN), usually an LSTM, and the selective attentive operator \emph{read}.

The \textit{read} operator is used to force the network to only see a certain portion of the original frame. Using the abstract representation encoded by the RNN, the parameters of a grid of 2D Gaussian filters are first computed and then used to extract a $N \times N$ patch of the image. The final patch is obtained through a fixed number of progressive refinements in which the RNN, starting from the whole frame at the beginning, progressively modifies its previous representation to better zoom on the salient portion of the image. Varying the stride and variance of the filters, the network can adaptively enlarge or reduce its field of view while still extracting patches of a fixed dimension.

More specifically, denoting as $\mathbf{h}_{t}$ the output of the RNN at the time $t$, a patch is extracted as it follows:
\begin{equation}
\label{eq:draw_read}
\mathit{read}(\mathbf{x}, \mathbf{h}_{t}^{dec}) =
\gamma\, (\mathbf{F}_Y\,\mathbf{x}\,\mathbf{F}_X^T )
\end{equation}
where $\mathbf{F}_Y$ and $\mathbf{F}_X^T$ (with dimension $N \times H_f$ and $W_f \times N$ respectively) are the Gaussian filters obtained using a linear transformation of $\mathbf{h}_{t}$, $x$ is the original $H_f \times W_f$ frame, and $\gamma$ is a scalar.

Even if originally designed for image generation, this procedure can also be used as an attention mechanism in object recognition architectures. Please refer to the original DRAW paper~\cite{Gregor2015Feb} for a detailed description of the model.

\vspace{-10pt}
\paragraph{Leaky Frame Integration}
\label{sec:frame_integration}

All the attention mechanisms designed to improve the pLSTM Recognition Network proposed in this paper are based on the frame reconstruction procedure described in \cite{Cannici2018May}. This simple mechanism, inspired by the functioning of spiking neurons, integrates events in time producing a sequence of frames on which conventional computer vision techniques can be applied. 
The pixel values of the reconstructed frame are updated whenever a new event $e = (x_e,\, y_e,\, ts)^t$ arrives, as it follows:
\begin{align}
\label{eq:integrator}
q_{x_m, y_m}^t &= max(p_{x_m, y_m}^{t-1} - \lambda \cdot \Delta_{ts}, 0) \\
p_{x_m, y_m}^t&=
\begin{cases}
 q_{x_m, y_m}^t + \Delta_{incr} 	&if (x_m, y_m)^t = (x_e, y_e)^t\\
 q_{x_m, y_m}^t 					&otherwise
\end{cases},
\end{align}
where $\Delta_{ts}=ts^t - ts^{t-1}$ decrements the whole frame of a quantity that depends on the time elapsed between the last received event, $ts^t$, and the previous one. As in the original YOLE paper~\cite{Cannici2018May}, we fix $\Delta_{incr}=1$, varying only $\lambda$ based on the dataset to be processed and in particular on the speed at which objects move. 

\section{Patch Extractor Recognition Network} \label{sec:patch_extractor_network;chap:models}

By monitoring the events activity inside the field of view of the neuromorphic camera, regions of interest can be detected and used as candidates for the object recognition process. For this purpose, we developed an algorithm that detects peaks of events activity and uses them to extract patches from reconstructed frames.
This approach takes inspiration from the spiking recognition network proposed in~\cite{Zhao2015Sep}, where a peak detection mechanism is used to decide when to output predictions. Instead of leaky \emph{integrate-and-fire} neurons, however, our method makes use of region-wise events statistics to identify and localize peaks.

\subsection{Peak Detection Algorithm} \label{sec:peak_detection_algo;chap:models}

The Peak Detection Algorithm we designed subdivides the $H_f \times W_f$ field of view into a grid of possibly overlapping $H_r \times W_r$ regions spaced by a fixed stride $s_r$. 
A moving window in time of length $L_{w}$ is associated to each tile; each \emph{activity value} of $L_{w}$ represents the number of events received inside the region within a certain interval of length $L_{bin}$. These \emph{activity windows} are used to detect peaks of activity inside each region by comparing the value contained in a fixed position $R_w$ of the window, which we call \emph{representative value}, with the remaining activity values in the same window.
As time passes, each activity value slides through the activity window and therefore, at some time, each value becomes the representative value $R_w$. We usually set $R_w$ to be the middle point in the window, but other configurations are also possible.

\begin{figure}
	\centering
	\includegraphics[width=0.76\linewidth]{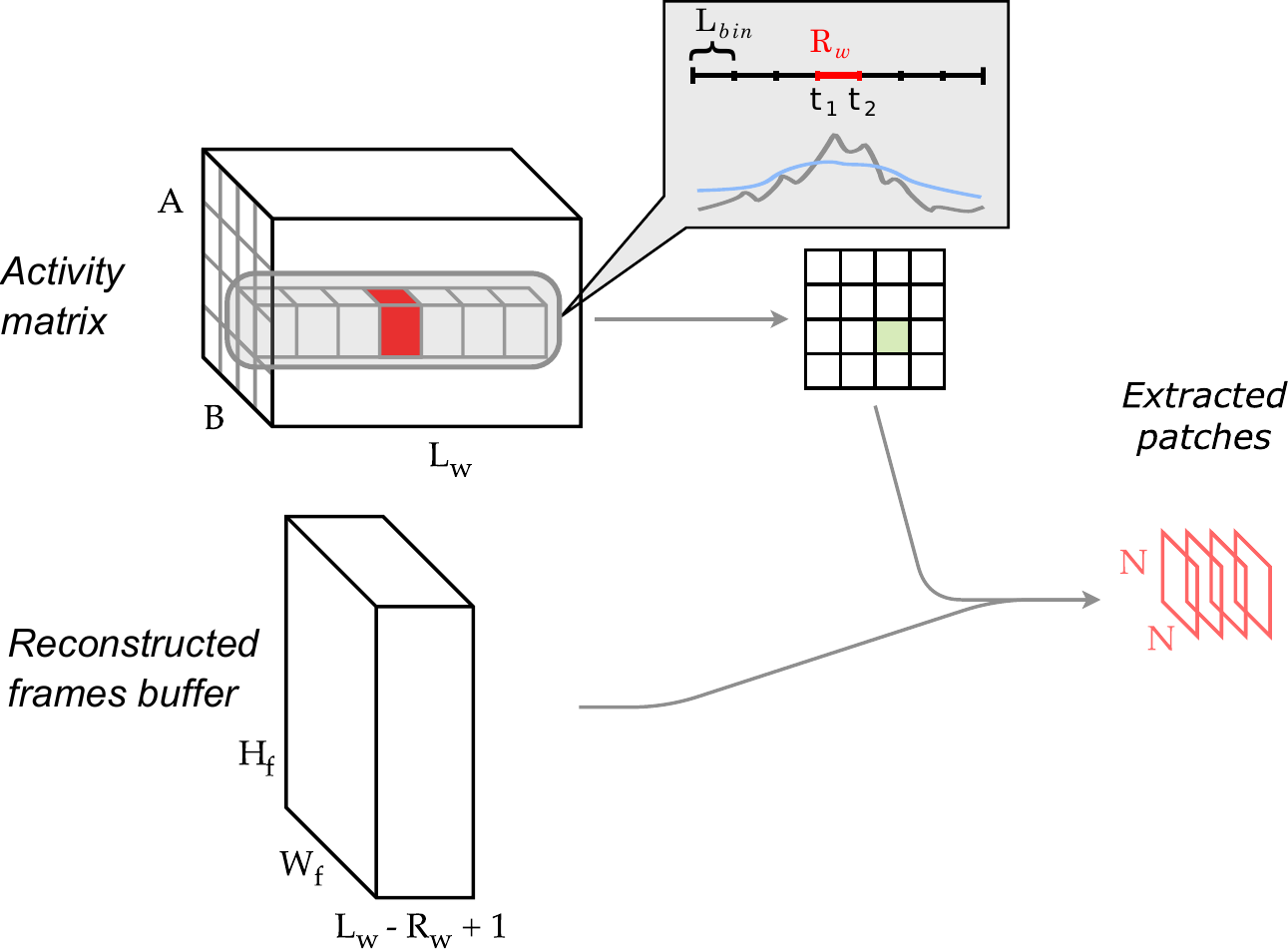}
    \vspace{-0.25cm}
	\caption[The peak detection process.]{The detection of a peak in position $R_w$ and the confidence interval, as a blue line, are represented at the top of the figure.
    For simplicity, windows are grouped together into an \emph{activity matrix}.}
	\label{fig:peak_detection}
    \vspace{-0.4cm}
\end{figure}

Periodically, each activity window is checked in order to determine the presence of peaks. A peak of activity is detected in a certain region whenever $R_w$ becomes the maximum value inside the window. In this case, the interval $(t_1, t_2)$, with $t_2 = t_1 + L_{bin}$, corresponding to the representative value is considered a peak and a \emph{patch extraction algorithm} is used to extract $N \times N$ patches inside this region using the frame reconstructed at the time instant $t_2$.

Since $R_w$ is usually not the first element of the window, the algorithm must wait the following $L_w - R_w + 1$ intervals before $(t_1, t_2)$ becomes the representative value and can consequently be analyzed. Being the peak detection delayed of $L_w - R_w + 1$ intervals, a buffer of integrated frames must be maintained to allow the extraction of patches from the right frame.

Every time a new event $\mathbf{e} = (x_e, y_e, ts_e)$ arrives, the frame in the most recent position of the buffer is updated, as described in Section \ref{sec:frame_integration}, as well as all the activity windows associated to regions in which $e$ is contained. Peak detection is only performed if the current interval has finished (\ie, $ts_e$ is more that $L_{bin}$ time instants after the beginning of the current interval) and at least $L_w$ activity values have been accumulated. In this case all the activity windows are checked and patches are extracted whenever a peak is detected. At the end of this process, the oldest frame in the buffer and the oldest values in all activity windows are removed to make room for the next interval.

To avoid false detections caused by noisy events received during time intervals of poor events activity and increase the robustness of the algorithm, we enhanced the peak detection procedure with a moving average approach.
A peak is considered to be valid if its value $x$ is above the confidence interval $x > \mu^t + \alpha \cdot \sigma^t$ where $\alpha$ is a parameter and $\mu^t$, $\sigma^t$ are the mean and standard deviation statistics of the whole field of view.

These are updated at the end of each interval as it follows:
\vspace{-0.2cm}
\begin{equation}
\label{eq:mean_std}
\mu^t = \frac{sum_{val}}{N_{val}}, \quad \sigma^t = \sqrt{\frac{sum_{{val}^2}}{N_{val}} - (\mu^t)^2}
\end{equation}
where ${sum_{val}}$ and $sum_{{val}^2}$ are respectively the sum of the activity values and the sum of their squares, and $N_{val} = N_{int} * A * B$, with $N_{int}$ the number of processed intervals. Both ${sum_{val}}$ and $sum_{{val}^2}$ are incrementally updated at the end of each interval. The equation of $\sigma^t$ is obtained from the relation between the mean and the variance of a stochastic variable, namely $Var[X] = \mathbb{E}[X^2] - {\mathbb{E}[X]}^2$.

\begin{figure}
	\centering
	\begin{subfigure}[c]{0.21\textwidth}
		\centering
		\includegraphics[width=\linewidth]{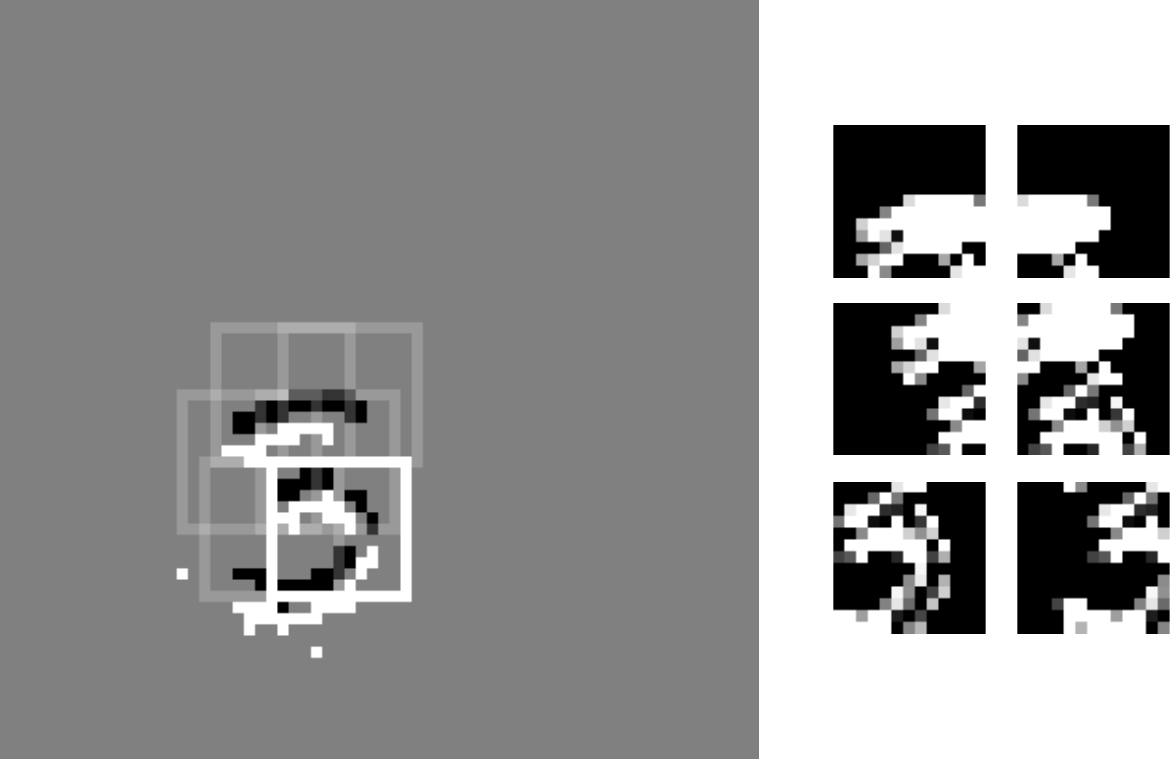}
		\caption{} \label{fig:patch_follower}
	\end{subfigure}%
	\hfill
	\begin{subfigure}[c]{0.21\textwidth}
		\centering
		\includegraphics[width=\linewidth]{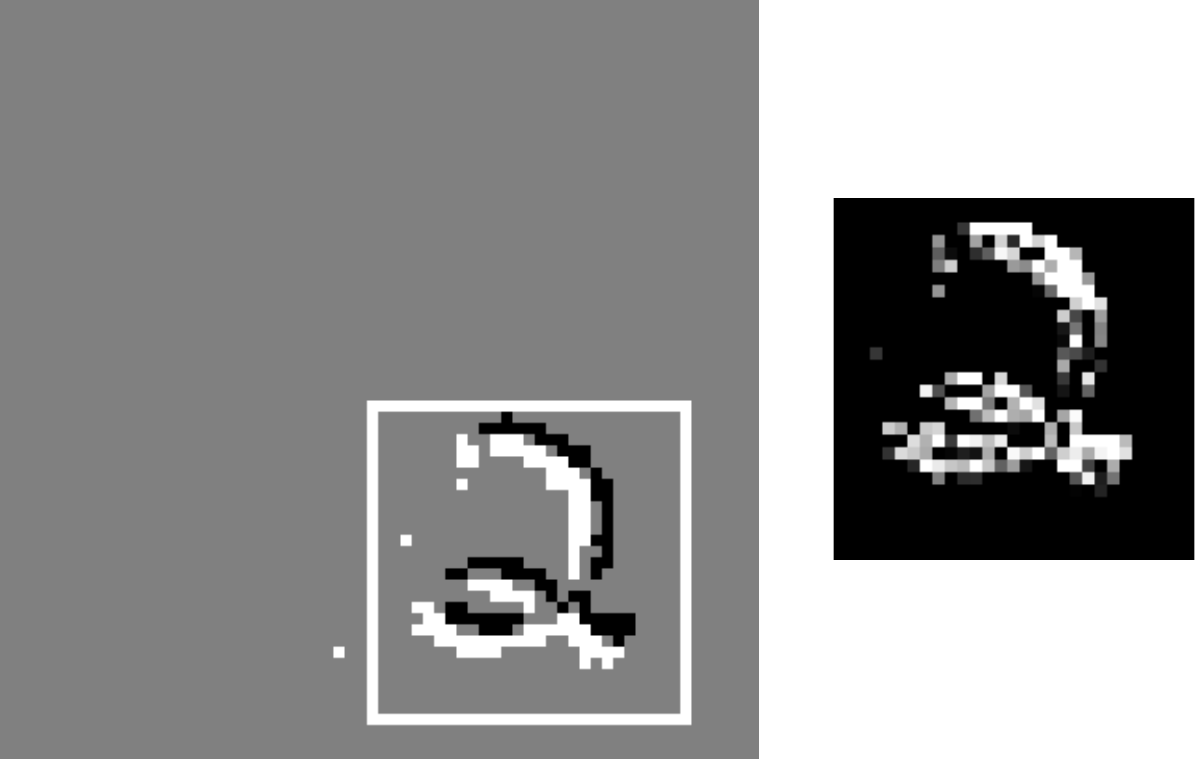}
		\caption{} \label{fig:patch_centered}
	\end{subfigure}%
    \vspace{-0.36cm}
	\caption[Comparison between patches extracted with the two variants of the patch extraction algorithm.]{Comparison between patches extracted with the two versions of the patch extraction algorithm. \textbf{(a)} The follower and \textbf{(b)} centered variants.}
	\label{fig:patch_comparison}
    \vspace{-0.4cm}
\end{figure}

\subsection{Patch Extraction Algorithms} \label{sec:patch_extraction}
We developed two mechanism for patches extraction. One that covers the whole object by centering a patch on activated regions, which we called \emph{Centered Patch Extraction}, and the other one which instead extracts small details by following the contours of the objects, which we called \emph{Follower Patch Extraction}. Examples of patches extracted with these two methods are shown in Figure \ref{fig:patch_comparison}. A video showing the detection of peaks and the extraction of patches on similar event-based recordings is available at \url{https://youtu.be/BV_ikdS4m3g}.

\paragraph*{Centered Patch Extraction}
The result provided by the peak detection unit is a two-dimensional boolean matrix that indicates which regions of the $A \times B$ grid activated, \ie, in which of these regions a peak has been detected. 
The goal of the Centered Patch Extraction algorithm is to extract patches which cover as much as possible the detected object. For this reason, active regions are grouped into \emph{macro-regions} by joining together adjacent active regions. For each macro-region one or multiple equally spaced $N \times N$ patches are extracted by covering the entire activated region.
This procedure is performed for every group of active regions and all the extracted patches are labeled with the timestamp associated to the frame from which they have been extracted.

\paragraph*{Follower Patch Extraction}
In the Follower variant of the Patch Extraction algorithm we choose the dimensions of the patches so that only small object details are extracted. We then extract patches by following the object outline with a simple recursive algorithm that extracts a patch as soon as an uncovered object pixel is visited.
As for the centered version of the algorithm, the timestamp of the frame from which patches have been obtained is also saved.

\begin{figure}[t!]
	\centering
	\includegraphics[width=\linewidth]{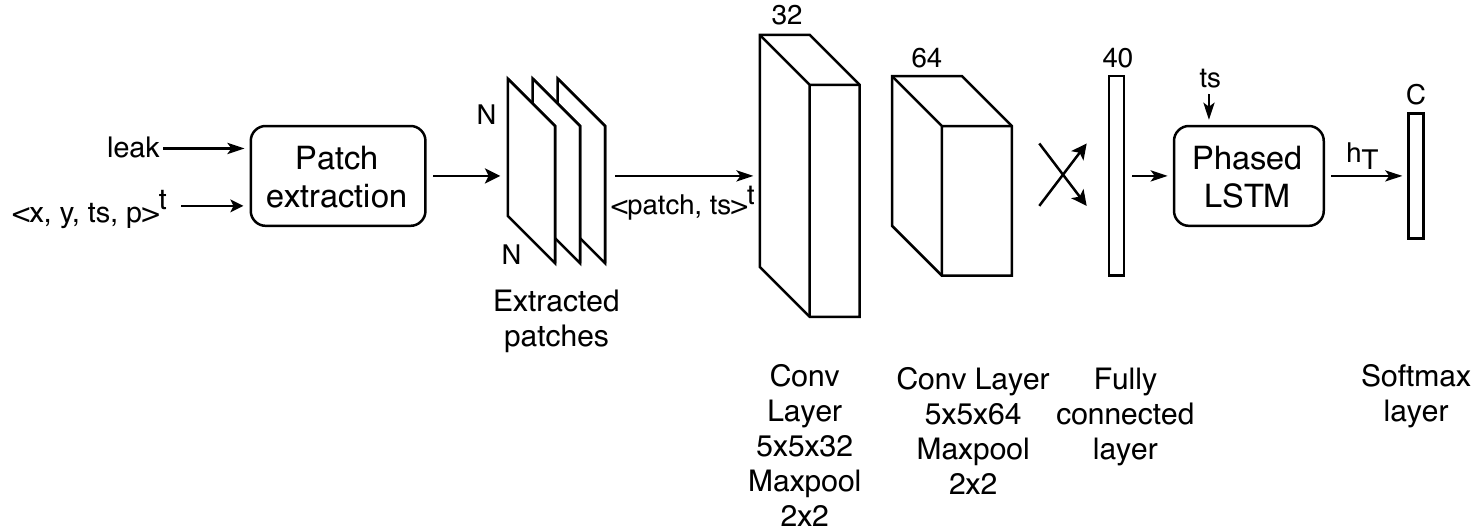}
    \vspace{-0.6cm}
	\caption[The patch extractor network for centered patches.]{The convolutional network used to classify the sequence of extracted patches.}
	\label{fig:patch_cnn}
    \vspace{-0.4cm}
\end{figure}

\subsection{Classification network}
The sequence of extracted patches constitutes the input of the recognition network that uses the timestamp information to correlate patches over time by means of a pLSTM layer (Figure \ref{fig:patch_cnn}).
The network is similar to the original pLSTM recognition network from \cite{Neil2016Oct}, where the word embedding layer has been replaced with a convolutional neural network. We used feature vectors of the same length of the ones extracted by the original embedding layer. However, no polarity information is added in this case.

The structure of this network is based on the idea that the patch extraction algorithm can be used as a way to convolve filters sparsely in space and time, driven by the events activity. Each extracted patch can indeed be considered as a single receptive field on which a small portion of a wider convolutional network, that potentially covers the whole input frame, is applied. The patch extraction algorithm, by monitoring the events activity during time and selecting the active receptive fields, allows to compute an event-based convolution of the filters only when and where a peak of activity has been detected. Features extracted from these receptive fields are then used by the pLSTM to reconstruct the global appearance of the object and its output is finally used for the overall prediction. We used the same network with both versions of the patch extraction algorithm.

\section{N-DRAW Recognition Network} \label{sec:draw_network;chap:models}

The patch extraction algorithm we presented in the previous section effectively extracts patches from integrated frames coming from neuromorphic cameras. The fact that patch extraction is driven by the events activity and that patches are computed and analyzed only when enough information has been accumulated, make the algorithm fit well in event-based scenarios. 

However, both patch extraction networks require the tuning of dataset-specific parameters and are not able to adapt to objects of variable dimension.
To improve the performance of this network, which still achieve better results than the pLSTM baseline when objects are not centered in the filed of view, we extended the patch extraction network obtaining a trainable procedure based on DRAW \cite{Gregor2015Feb} and whose functioning is similar to the original procedure. Being designed on top of a recurrent neural network that gradually encodes visual information and being able to gradually adjust its predictions over time, DRAW naturally fits the sequential nature of event-based imaging. We finally used the networks presented in the previous sections as additional baselines to evaluate how much the network performance improves when a patch extraction procedure able to automatically adjust to changes is used.

We designed the \emph{N-DRAW patch-based} network by combining the architecture of the previous patch extraction algorithm with the DRAW recognition model. 
Then, we designed a second variant, \ie, the \emph{N-DRAW event-based} network, that directly uses the sequence of events as input to the encoder network.

\subsection{Patch-based model description (p-N-DRAW)} \label{sec:draw_patch_network;chap:models}

\begin{figure}
	\centering
	\includegraphics[width=\linewidth]{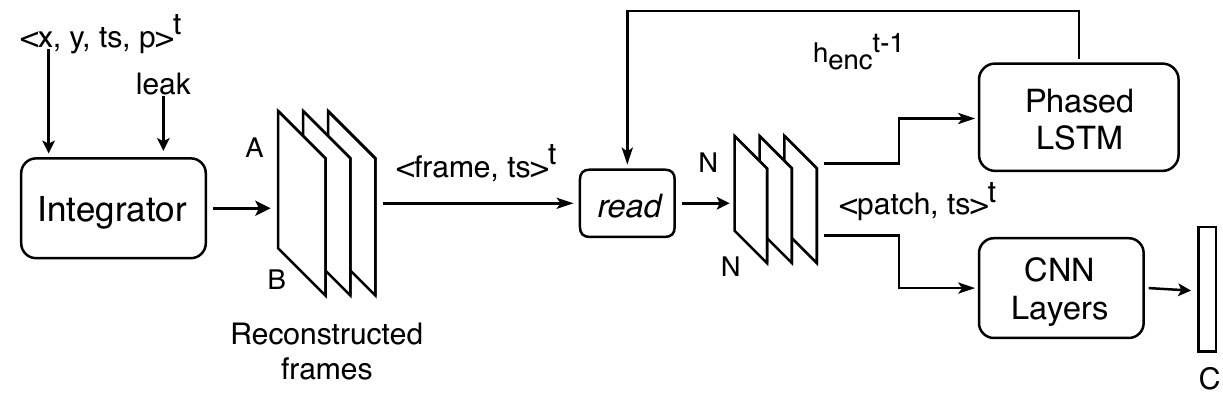}
    \vspace{-0.6cm}
	\caption{The p-N-DRAW recognition network.}
	\label{fig:draw_v3}
    \vspace{-0.3cm}
\end{figure}

We modified the original DRAW network to detect objects captured with event-based cameras by using a frame reconstruction mechanism as the first layer of our architecture (Figure \ref{fig:draw_v3}). The \textit{read} operation takes as inputs the most recent frame $frame^t$ and the output of the encoder at the previous time step $\mathbf{h}_{enc}^{t-1}$. This extracts the parameters of 2D Gaussian filters and uses them to transform the $A \times B$ input frame into a fixed size $N \times N$ patch. The timestamp $ts^t$ associated with the current frame is used as an additional input for the recurrent network. In contrast to the original architecture that uses a simple LSTM network, we use a pLSTM layer as the encoder so that the timestamp associated with each patch can also be exploited. By doing that, the network learns to sparsely update its internal representation based on the timing of the input features.

Differently from the original model, where the image is static, we deal instead with a sequence of integrated frames that may slightly differ from each other. The \textit{read} operation, therefore, has to decide where to attend at the current time step by using the encoder output produced while observing the previous frame, where the object may be in a slightly different position. We found, however, that this does not constitute a problem for the recurrent architecture since it can learn to compensate the objects movement by comparing consecutive frames.

We want our network to be able to recognize objects as soon as enough information has been accumulated. For this reason, we decided to perform a prediction regularly rather than after having seen the whole sequence, as opposed to the standard DRAW architecture. Since using every patch for prediction may prevent the network to learn a good extraction mechanism (not having the encoder a fixed reasoning period which can be used to gradually zoom and refine the prediction), we perform instead a prediction every $M$ successive frames.

\begin{figure}
	\centering
	\includegraphics[width=\linewidth]{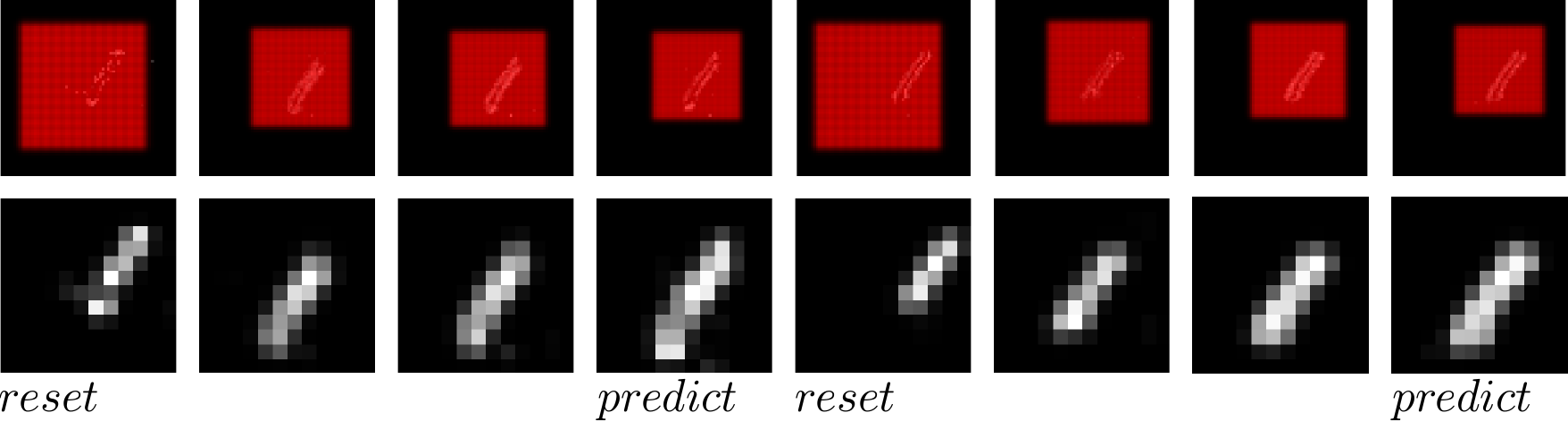}
    \vspace{-0.5cm}
	\caption[DRAW's attention mechanism applied to event-based data.]{The top row shows $8$ consecutive $68 \times 68$ frames and, in red, the grid of $12 \times 12$ 2D Gaussian filters, whereas the bottom one the corresponding extracted patches. In this example, the encoder state is reset every $4$ frames, which causes the filter to reposition itself to cover most of the frame.}
	\label{fig:draw_every_4}
    \vspace{-0.4cm}
\end{figure}

If $M$ is not too large (we used $M = 4$ in our experiments), the network can still generate predictions quite often allowing the model to be used for continuous classification. After the fixed $M$ steps, the internal state of the encoder can either be reset or maintained as a starting point for the next prediction. We found to be beneficial to maintain the internal state when objects do not move too much, as the network can continue to refine the previous prediction. However, if objects move fast the network performs better when the state is reset, as this allows it to see the whole frame and progressively re-locate the object. An example is shown in Figure \ref{fig:draw_every_4}. 

\subsection{Event-based model description (e-N-DRAW)} \label{sec:draw_event_network;chap:models}

\begin{figure}
	\centering
	\includegraphics[width=\linewidth]{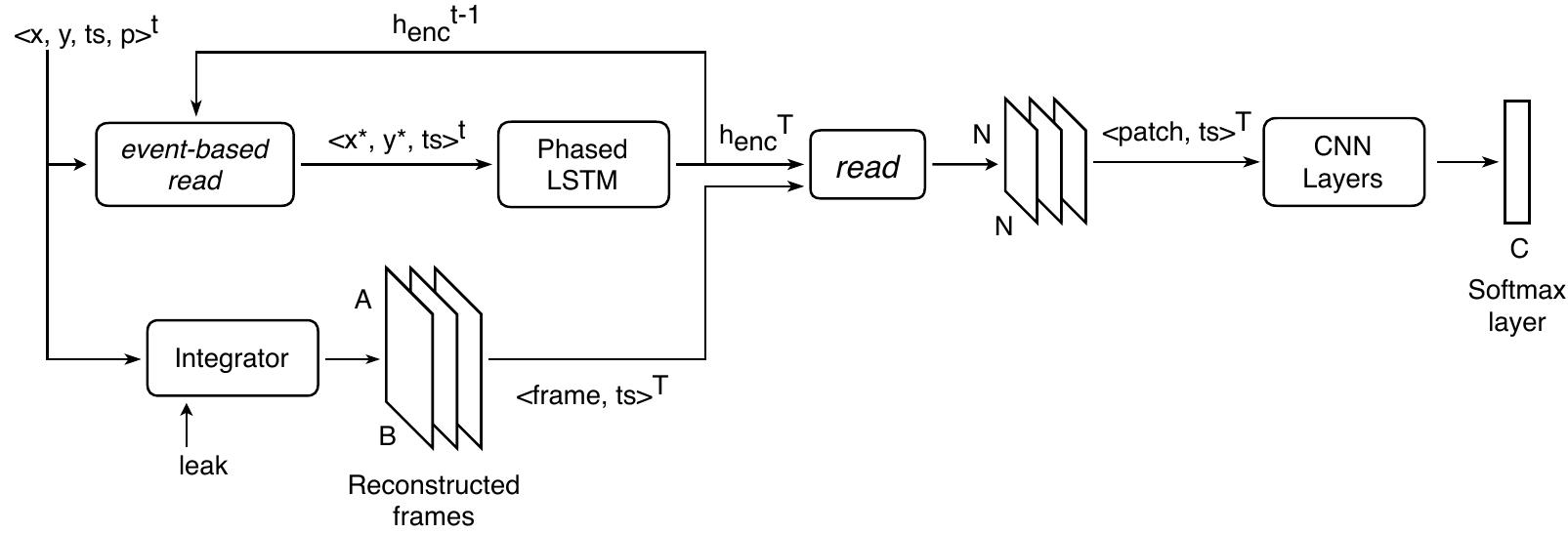}
    \vspace{-0.4cm}
	\caption{The e-N-DRAW architecture. The projected coordinates $(x_e^*, y_e^*)$ are used to guide the attention mechanism in finding the filters parameters (loop connection at the top of the figure).}
	\label{fig:draw_v4}
    \vspace{-0.4cm}
\end{figure}

\textit{N-DRAW event-based} extends the DRAW attention mechanism to directly process the stream of incoming events and uses it as a reference to locate the relevant part of the scene, in a similar way as in the patch extraction algorithm.
This variant, depicted in Figure \ref{fig:draw_v4},  makes use of two read operations: \emph{event-based read}, the modified attention mechanism that processes events, and \emph{read}, the vanilla DRAW's operation that extracts patches from reconstructed frames. The sequence of events is partitioned into \emph{intervals} of equal temporal length $T$. Events are used both to reconstruct frames through the frame integration procedure and to detect the relevant part of the scene by means of the recurrent pLSTM layer. Once the whole sequence of events inside the interval has been processed by the encoder, its output $\mathbf{h}_{enc}^T$, is used to extract a $N \times N$ patch from the last integrated frame $frame^T$ using the standard extraction procedure \textit{read}. The extracted patch is then processed as usual by applying a sequence of convolutional layers and by using the extracted representation to predict the class label.

The \textit{event-based read} projects the input coordinates into the \emph{patch space} to provide the encoder network a feedback on the transformation applied by the Gaussian filters. Given an event at location $(x_e, y_e)$ in the input space, the \textit{event-based read} produces as output a new event with the same timestamp $ts$, but having as coordinates $(x_e^*, y_e^*)$ those that the original event has in the $N \times N$ patch space. In this way the encoder network progressively follows the events activity and modifies its internal state so that the extracted patch will be centered on the object.

The operation performed by the \textit{event-based read} procedure is based on the original transformation. A frame containing a single positive pixel in correspondence of the incoming event coordinates $(x_e, y_e)$ is considered. A patch is extracted from this frame by means of the original \textit{read} operation using the set of Gaussian filters obtained from the encoder output at the previous step $\mathbf{h}_{enc}^{t-1}$. This patch contains a possibly blurred dot in a certain location whose coordinates $(x_e^*, y_e^*)$ can be obtained by looking at the brightest pixel of the patch. In particular, the output coordinates are defined as $(x_e^*, y_e^*) = \argmax_{(x,y)} \gamma \left( \mathbf{F}_Y\,\mathbf{x}\,\mathbf{F}_X^T \right)$, where $\mathbf{F}_Y$ and $\mathbf{F}_X$ are the set of extracted Gaussian filters. The \textit{event-based read} and \textit{read} operations share the same linear transformation that allows the encoder output to be transformed into the filter parameters. In this way, the same transformation learned while observing the sequence of events will be also used to extract the actual patch from the reconstructed frame.

\begin{figure}
	\centering
	\includegraphics[width=\linewidth]{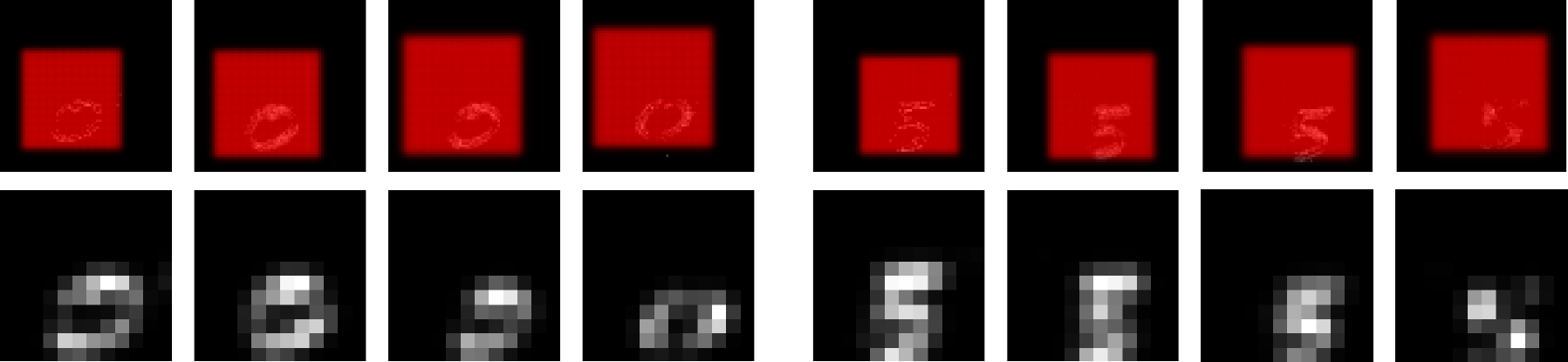}
    \vspace{-0.4cm}
	\caption[Patches extracted with the event-based DRAW architecture.]{The filter changes during the successive stages of the patch extraction process on two Shifted N-MNSIT examples. Contrary to Figure \ref{fig:draw_every_4}, we do not see the filter gradually zooming in, since in this case the network uses the sequence of events to progressively refine its prediction. When the filter is first applied to the frame, it has already been perfected.}
	\label{fig:draw_event}
    \vspace{-0.2cm}
\end{figure}

As it happens with the original attention mechanism, the network starts by considering a patch that roughly covers the whole frame so that most of the incoming events will be contained in the extracted patch. As more events arrive, the network becomes more confident about the position of the object and it starts reducing the dimensions of the filter ignoring irrelevant events. Events that are not contained inside the filter's region, \ie, those for which the patch extracted with the \textit{event-based read} operation is completely blank, are ignored by the network that skips them during the recursive execution. This is in line with the original DRAW patch extraction procedure which ignore the part of the scene not contained inside the extracted patch.
Figure \ref{fig:draw_event} shows the successive stages of the \emph{event-based} patch extraction procedure on some Shifted N-MNIST examples.

This attention procedure resemble the patch extraction algorithm we presented in the previous section; the spatial location of the events is directly used to discover regions of interest in the input scene which are then used to extract patches from integrated frames. This new mechanism has the advantage of being a trainable procedure that can be learned together with the rest of the classification network thanks to its fully-differentiable nature. As for the patch extraction algorithm, though, this mechanism only bases its predictions on the events activity, without any visual feedback, as opposed to the \emph{patch-based} architecture. This characteristic limits the network performance in scenarios in which the background is also moving with respect to the camera and the attention mechanism has to discriminate between events emitted by the object and those emitted by the rest of the scene, as we registered testing the network on CIFAR10-DVS \cite{Li2017May} and N-Caltech101 \cite{Orchard2015Nov} datasets. Moreover, we found that this network has difficulties in centering and zooming on the object with respect to the \emph{patch-based} one, as reported in the next section.

\section{Experiments}
\paragraph{Datasets}
The performance of the proposed attention-based networks have been tested on four datasets available in literature, the N-MNIST \cite{Orchard2015Nov}, MNIST-DVS \cite{Serrano-Gotarredona2015Dec}, CIFAR10-DVS (CIF10) \cite{Li2017May} and N-Caltech101 (Cal101) \cite{Orchard2015Nov} collections. These datasets have been obtained by displaying original images in front of a neuromorphic camera and by moving them, or the camera itself, following a predefined trajectory that resemble human saccades. Since these MNIST datasets are quite simple, especially N-MNIST, we also considered the \emph{Shifted N-MNIST} (S-N) and \emph{Shifted MNIST-DVS} (S-DVS) variants \cite{Cannici2018May} in which the original digits (\ie, the sequence of events representing them) are placed in a random location of a bigger field of view.

\subsection{Experiments Setup}

\begin{table}[t]
	\caption{Patch extraction algorithm parameters.}
	\label{tab:patch_extr_params}
    \vspace{-0.2cm}
	\centering
	\scriptsize
    \newcolumntype{a}{>{\centering\arraybackslash}p{0.6cm}}
    \newcolumntype{b}{>{\centering\arraybackslash}p{0.7cm}}
    \newcolumntype{M}{>{\centering\arraybackslash}p{0.3cm}}
    \newcolumntype{N}{>{\centering\arraybackslash}p{0.4cm}}
	\begin{tabular*}{\linewidth}{abMMMMMNNN}
    	\toprule
    	& &  \multicolumn{5}{c}{S-DVS} & S-N & CIF10 & Cal101 \\
		& & sc4 & sc8 & sc16 & sc4+8 & all \\
		\midrule
        \multirow{3}{*}{centered}
          & $s_r$ 		& 11  & 24  & 24  & 24  & 24  & 5  & 10 & 10 \\
          & $W_r$=$H_r$ & 24  & 32  & 32  & 32  & 32  & 23 & 48 & 48 \\
          & $N$ 		& 29  & 55  & 105 & 55  & 105 & 29 & 105 & 105 \\
        \midrule
        \multirow{3}{*}{follower}
          & $s_r$ 		& 5   & 15  & 24  & 24  & 24  & 5 & 12 & 12 \\
          & $W_r$=$H_r$ & 9   & 23  & 32  & 32  & 32  & 9 & 32 & 32 \\
          & $N$ 		& 13  & 23  & 53  & 23  & 53  & 13 & 75 & 75 \\
		\bottomrule
	\end{tabular*}
    \vspace{-0.4cm}
\end{table}

All the results presented in this paper were obtained by optimizing the cross entropy loss function using Adam \cite{Kingma2014Dec} with default parameters ($\beta_1 = 0.9$, $\beta_2 = 0.999$, $\epsilon=10^{-8}$) and learning rate $10^{-4}$. Networks parameters were initialized using the mechanism proposed in \cite{Glorot2010Mar} and early-stopping was applied to prevent overfitting.

\paragraph*{Patch extractor networks}

Due to the high number of parameters on which the patch extraction algorithm depends, we decided to fix part of their values manually by inspecting the quality of the results produced by the patch extraction process, \ie, the sequence of extracted patches. Even though this approach does not provide a complete exploration of the space of possible values and does not allow to compare the recognition performances of the final model in response to the change of every single parameter, it enabled us to quickly analyze their effects and determine the way these parameters interact with each other.

The parameters of the activity windows were chosen by analyzing the rate of events generated from the event-based camera during the entire recording period. As reported in~\cite{Orchard2015Nov}, N-MNIST peaks of activity are correlated with the speed of the objects moving inside the scene. Since the same set of movements has been used to record the entire dataset, the activity outlines are very similar between examples. For these reasons we used $L_w = 101$, $R_w = 51$ and $L_{bin}$ = $1$ms, obtaining an activity window that covers $101$ms, which is roughly the temporal length of the saccade movement used to record digits. We used the same parameters also with N-Caltech101 samples, as they have been recorded with the same procedure, and with CIFAR10-DVS recordings. MNIST-DVS digits, instead, being registered using a different and more noisy procedure, do not show a clear activity outline. To cope with the higher variability of event peaks and make a more reactive detection, we decided to use a smaller window with parameters $L_w = 81$, $R_w = 41$, and the same interval length $L_{bin} = 1$ms.

Regions parameters were chosen in order to cover a good portion of the objects and also to make regions overlap between each other to obtain good translation invariance properties.
The patch dimension $N \times N$ was chosen in such a way to extract the entire object with the \emph{centered} patch extraction procedure, and to cover only small details of each object in the \emph{follower} variant.
Table \ref{tab:patch_extr_params} reports the specific parameters we used.

\paragraph*{N-DRAW networks}

The hyperparameters for the N-DRAW architectures were chosen using a ``greedy'' approach for parameter optimization. We focused on the optimization of a single parameter at the time by gradually changing its value and registering the resulting effect in network performance.

Since N-MNIST and scale4 (sc4) MNIST-DVS digits were recorded to have roughly the same dimensions of the original MNIST digits, we decided to use the same patch size, \ie, $12 \times 12$, that was used with the original DRAW architecture \cite{Gregor2015Feb} to process Cluttered MNIST \cite{Mnih2014Jun} digits, a variation of MNIST in which digits are placed in a random location of a bigger frame, similarly to the Shifted N-MNIST and Shifted MNIST-DVS collections. Given that digits of a certain MNIST-DVS scale are roughly double the size of the previous scale, we used $24 \times 24$ and $48 \times 48$ patches for \emph{scale8} (sc8) and \emph{scale16} (sc16) examples. We finally used $48 \times 48$ patches also for N-Caltech101 and CIFAR10-DVS samples since they both feature objects that occupy most of the $128 \times 128$ frame, similarly to \emph{scale16} MNIST-DVS samples.

The number $M$ of recursive iterations was instead determined by using a simplified version of the network, which resembles the original DRAW classification network.
We found $M = 4$ to be the optimal value.

Finally, to process the extracted patches, we used the same set of convolutional layers we used in the patch extractor architectures, \ie, two convolutional layers and a fully connected layer that maps features into $40$-dimensional vectors. The size of the encoder network was set to be equal to the number of cells used in the original pLSTM recognition network, \ie, 110.

\subsection{Results and Discussion}

\paragraph{Baseline}

\begin{table}[t]
	\caption[pLSTM network's accuracies on the Shifted N-MNIST]{pLSTM's baseline accuracy on the Shifted N-MNIST.}
    \vspace{-0.25cm}
	\label{tab:phasedlstm_shifted_nmnist}
	\centering
	\scriptsize
	\begin{tabular}{ccccc}
		\toprule
		Frame	& Embedding	& Encoder	&	Augmented & Test Accuracy \\
		\midrule
		$34 \times 34$ (original)			& 41				& 110			&	No					& 97.4 \\
		$68 \times 68$			& 41				& 110			&	No					& 26.0 \\
		$68 \times 68$			& 101				& 200			&	No					& 81.7 \\
		$68 \times 68$			& 101				& 200			&	Yes					& 90.3 \\
		\bottomrule
	\end{tabular}
    \vspace{-0.1cm}
\end{table}

We compared the performance of the proposed models with the results obtained by the Phased LSTM object recognition network described in \cite{Neil2016Oct}. All the proposed networks are indeed based on pLSTM cells and they were originally designed to overcome some limitations of the original pLSTM model. Since the pLSTM architecture only uses an embedding layer to extract features from events, it does not show any scale or translation invariance property, as reported in Table \ref{tab:phasedlstm_shifted_nmnist}. The loss in performance reduces when an augmented Shifted N-MNIST dataset, obtained randomizing the position of each digit after every epoch, and therefore using a higher number of training samples, is used to train the model.

Table \ref{tab:mnst-dvs_multiscale_comparison} reports the results obtained on the Shifted N-MNIST, Shifted MNIST-DVS, CIFAR10-DVS and N-Caltech101 datasets, using the same layers configuration described in \cite{Neil2016Oct}. To reduce the size of the model (which depends on the frame size due to the presence of the embedding layer) and speed up training, we cropped the central portion of MNIST-DVS examples obtaining smaller samples containing only the digits. In particular we used $35 \times 35$, $65 \times 65$ and $100 \times 100$ field of views for the sc4, sc8 and sc16 examples respectively. In case of mixed scales we used the size of the bigger scale in the dataset.

\paragraph*{Results}

\begin{table}[t]
	\caption{Comparison between the performances of the proposed models.}
    \vspace{-0.25cm}
	\label{tab:mnst-dvs_multiscale_comparison}
	\centering
	\scriptsize
    \newcolumntype{S}{>{\centering\arraybackslash}m{0.35cm}}
    \newcolumntype{C}{>{\centering\arraybackslash}m{0.45cm}}
    \newcolumntype{N}{>{\centering\arraybackslash}m{1.2cm}}
	\begin{tabular*}{\linewidth}{NSSSSSCCC}
		\toprule
        			& \multicolumn{5}{c}{S-DVS} & S-N & CIF10 & Cal101 \\
					& sc4 & sc8 & sc16 & sc4+8 & all \\
		\midrule 
		pLSTM  & 82.20 & 87.01 & 81.60 & 86.60 & 83.63 & 90.30 & 17.10 & 1.39 \\
        \midrule 
		p. centered  & \textbf{98.30} & 95.90 & 96.30 & 95.90 & 95.53 & \textbf{97.37} & \textbf{44.10} & 21.39 \\
        p. follower & 91.30 & 90.50	& 95.10 & -	& - & 91.07 & 37.40 & 18.47 \\
		e-N-DRAW & 91.35 & 96.50 & 95.69 & 96.74 & 95.10 & 92.30 & 36.89 & \textbf{28.95} \\
        p-N-DRAW & 94.81 & 96.88 & 95.32 & \textbf{97.96} & 93.19 & 96.42 & 38.17 & 27.69 \\
		p-N-DRAW (reset) & 94.10 & \textbf{97.39} & \textbf{96.71} & 96.61 & \textbf{98.24} & 95.15 & 41.29 & 27.70 \\
		\bottomrule
	\end{tabular*}
    \vspace{-0.35cm}
\end{table}

Table \ref{tab:mnst-dvs_multiscale_comparison} shows the results we obtained on multiple datasets using the proposed models. All the models achieve better results w.r.t. the pLSTM architecture, highlighting the advantages of using attention mechanisms to improve translation invariance.

As expected, the \emph{follower} variant of the patch extraction network achieved worse results with respect to the \emph{centered} version. While the use of smaller patches allows the network to maintain its event-based nature, reacting to small details as soon as they become visible, the task the pLSTM layer needs to learn is much harder. The overall appearance of the object needs indeed to be reconstructed by only looking at the sequence of details, whose order is not always the same among objects of the same class since it depends on where and when peaks are detected. Note that we did not test the follower patch-extraction algorithm on mixed MNIST-DVS scales because, being patches of fixed dimensions, this would have meant to extract small details in sc16 samples, or the entire digit in sc4 samples.

The N-DRAW patch-based variant performed better than its event-based counterpart in almost all datasets. This difference in accuracy is explained by the fact that the encoder of the event-based architecture predicts the final set of filters parameters only based on the sequence of incoming events. This mechanism does not provide any visual cue regarding the effect that the set of predicted filters have on the actual extracted patch.
However, even if it does not reach the same classification accuracies of the N-DRAW patch based algorithm, this mechanism still represents a valid event-based attention mechanism being able, by only using the events sequence, to identify regions of interests inside the scene. Using this procedure we obtained indeed similar results of the ones achieved using the centered patch extraction network in almost all datasets. 

\begin{figure}
	\centering
	\includegraphics[width=0.95\linewidth]{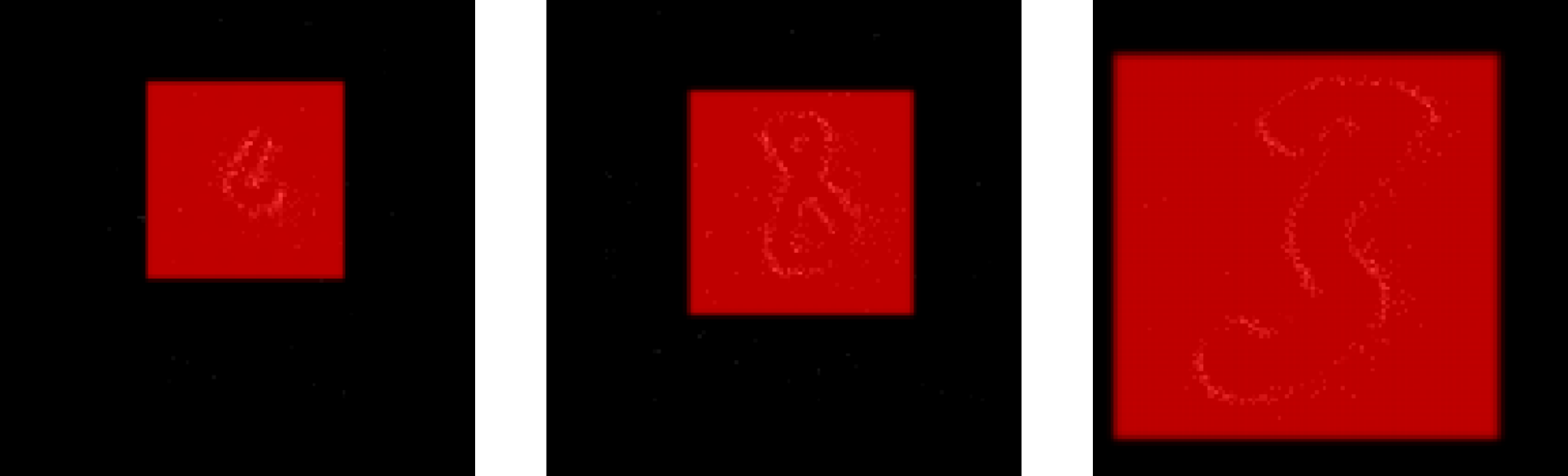}
    \vspace{-0.2cm}
	\caption[DRAW patch extraction procedure on MNIST-DVS multiple scales.]{The DRAW attention procedure learned by a single network trained to recognize all MNIST-DVS scales.}
	\label{fig:draw_filter_allscales}
    \vspace{-0.4cm}
\end{figure}

When evaluated on datasets composed of multiple scales, the N-DRAW architecture outperforms the patch extractor network but using a fully trainable model. 
N-DRAW allows indeed to adaptively zoom on the objects and enables the extraction of patches containing a reduced variability of objects dimensions. Large objects are extracted as they are whereas smaller ones are enlarged to better fit the patch. This behavior is depicted in Figure \ref{fig:draw_filter_allscales}.

We also tested the proposed models on more challenging datasets in terms of both background noise and objects complexity, N-Caltech101 and CIFAR10-DVS. All the proposed networks achieve better results than the pLSTM baseline, showing advantages on the use of convolutional layers and attention mechanisms. The obtained results, however, do not achieve the state of the art accuracy obtained using the DART \cite{Ramesh2017Oct} descriptor ($65.43 \pm 0.35\%$ on CIFAR10-DVS and $65.6\%$ on N-Caltech101).
This lack of performance can be traced back to the poor capabilities of the proposed extraction algorithm to distinguish between background and foreground events in very noisy environments (where the moving average approach is not enough) and to the need of a greater number of samples to allow the trainable DRAW mechanism to learn effective filter transformations. When evaluating models on the two original image based collections, the lack of training samples is indeed usually addressed using pre-trained feature extractors obtained from larger collection of samples, which however are still missing in the neuromorphic field.

\section{Conclusions and Future Works}

In this paper we proposed two approaches for event-based visual attention. The first one makes use of a simple algorithm to identify regions of interest from events while improving the translation invariance properties of the original pLSTM model. The second one is a fully-differentiable procedure based on the popular DRAW attention mechanism which improves the scale invariance properties of the first network.
Using the proposed methods we were able to obtain promising results in improving the effectiveness of conventional CNNs for event-based computation obtaining an architecture capable to deal with real-world applications where it is likely to find objects in different positions and scales.

As a first improvement of the proposed models, we plan to extend the original leaky frame integration procedure with an adaptive procedure able to dynamically vary the leak parameter and adapt the trained model to the speed of observed objects.
Moreover, as we aim to design a fully event-based network which does not rely on reconstructed frames to recognize objects, we are also considering to extend the event-based N-DRAW model by directly processing the filtered coordinates with an additional pLSTM layer, as in the original Phased LSTM recognition network \cite{Neil2016Oct}, without making use of frames to extract features. The network could indeed still maintain good translation and scale invariance properties by exploiting the ability of its \emph{event-based read} to filter out irrelevant events while maintaining and centering the relevant ones in the network field of view.

As a final remark let's note that the feasibility of the proposed approach is also supported by the features of advanced many-core architectures, e.g., high-performance and low-latency communication subsystems \cite{cutbuf} and caches able to dynamically adapt their memory capacity \cite{zoni2018darkcache}, that allow to efficiently process the asynchronous and bursty sequence of events imposed by the event-based cameras.

{\small
\paragraph*{Acknowledgements}
The research leading to these results has received funding from project TEINVEIN: TEcnologie INnovative per i VEicoli Intelligenti, CUP (Codice Unico Progetto - Unique Project Code): E96D17000110009 - Call ``Accordi per la Ricerca e l’Innovazione", cofunded by POR FESR 2014-2020 (Programma Operativo Regionale, Fondo Europeo di Sviluppo Regionale – Regional Operational Programme, European Regional Development Fund).

\bibliographystyle{ieee}
\bibliography{Bibliography}}

\end{document}